%% file: main.tex
\documentclass[10pt,twocolumn,letterpaper]{article}

\usepackage{cvpr}              

\usepackage{graphicx}
\usepackage{amsmath}
\usepackage{amssymb}
\usepackage{booktabs}
\usepackage{mathtools}
\usepackage{iac_pkg}
\usepackage{lipsum}
\usepackage{amssymb}
\usepackage{pifont}
\usepackage{multirow}
\usepackage{tabularx, booktabs}
\usepackage[htt]{hyphenat}
\usepackage{color}
\usepackage{xspace}
\usepackage{cite}
\usepackage{overpic}
\usepackage{arydshln}
\usepackage{subcaption}
\usepackage{xcolor}
\usepackage{bm}

\usepackage{cuted}
\usepackage{afterpage}

\definecolor{citecolor}{RGB}{34,139,34}
\definecolor{lightred}{RGB}{255,100,100}
\definecolor{cell_bisque}{rgb}{1.0, 0.89, 0.77}
\definecolor{cell_blond}{rgb}{0.98, 0.94, 0.75}
\definecolor{cell_blue}{RGB}{155, 187, 228}
\definecolor{princetonorange}{rgb}{1.0, 0.56, 0.0}
\definecolor{pinkpearl}{rgb}{0.91, 0.67, 0.81}
\definecolor{mossgreen}{rgb}{0.68, 0.87, 0.68}

\newcommand{\Paragraph}[1]{\vspace{-0mm}\noindent\textbf{#1.}\hspace{0mm}}
\newcommand{\Section}[1]{\vspace{-1mm} \section{#1} \vspace{-0mm}}
\newcommand{\SubSection}[1]{\vspace{-1mm} \subsection{#1} \vspace{-0mm}}

\usepackage{hyperref}
\hypersetup{pagebackref=true,breaklinks=true,letterpaper=true,colorlinks,bookmarks=false,citecolor=cyan}

\usepackage[capitalize]{cleveref}

\def\paperID{52} 
\def\confName{CVPR}
\def\confYear{2024}

\begin{document}
\title{ShadowRefiner: Towards Mask-free Shadow Removal via Fast Fourier Transformer}
\author{Wei Dong$^1$\quad Han Zhou$^{1*}$\quad Yuqiong Tian$^1$\quad Jingke Sun$^1$\quad Xiaohong Liu$^2$\quad Guangtao Zhai$^2$\quad Jun Chen$^{1*}$\\
$^1$McMaster University\quad $^2$Shanghai Jiao Tong University\\
{\tt\small wdong1745376@gmail.com}, {\tt\small \{zhouh115, tiany86, sun409\}@mcmaster.ca}\\ {\tt\small \{xiaohongliu, zhaiguangtao\}@sjtu.edu.cn}, {\tt\small chenjun@mcmaster.ca}
}


\twocolumn[{
\maketitle
\vspace{-8mm}
\begin{center}
    
    \captionsetup{type=figure}
    \setlength{\abovecaptionskip}{2mm}
    \centering
    \includegraphics[width=1\textwidth]{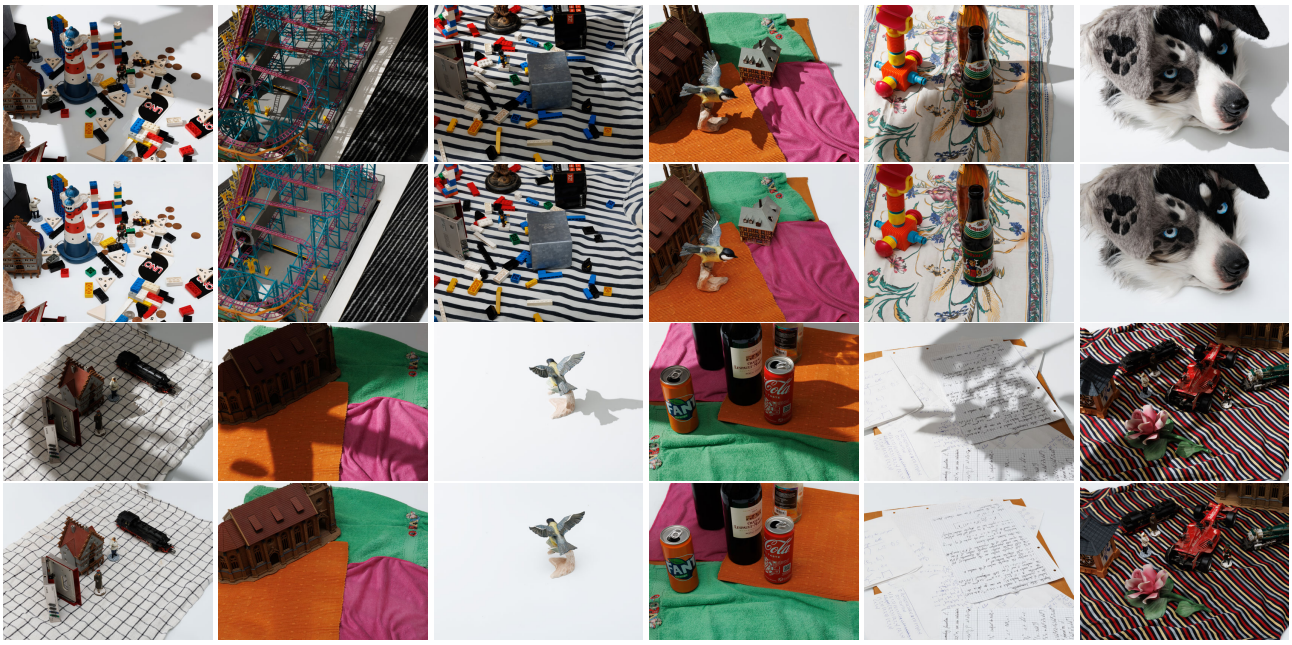}
    \captionof{figure}{Our test results on NTIRE 2024 Image Shadow Removal Challenge~\cite{vasluianu2024ntire_isr}. Our proposed method is the \red{\textbf{champion}} in the \red{\textbf{Perceptual Track}} and achieves the \blue{second best} performance in the \blue{Fidelity Track}.}
    \label{firstfigure}
\vspace{-1mm}
\end{center}
}]

\maketitle

\newcommand\blfootnote[1]{%
  \begingroup
  \renewcommand\thefootnote{}\footnote{#1}%
  \addtocounter{footnote}{-1}%
  \endgroup
}

\blfootnote{$^*$ Han Zhou and Jun Chen are corresponding authors}

\input{section/0_abstract}

\input{section/1_intro}
\input{section/figure/architecture}
\input{section/2_related}

\input{section/3_methods}

\input{section/4_experiments}

\input{section/4_experiment_ablation}
\input{section/5_conclusions}


\clearpage
{
    \small
    \bibliographystyle{ieeenat_fullname}
    \bibliography{main}
}

\end{document}

%% file: section/0_abstract.tex
\begin{abstract}
\vspace{-2mm}
Shadow-affected images often exhibit pronounced spatial discrepancies in color and illumination, consequently degrading various vision applications including object detection and segmentation systems. To effectively eliminate shadows in real-world images while preserving intricate details and producing visually compelling outcomes, we introduce a mask-free \textbf{Shadow} Removal and \textbf{Refine}ment netwo\textbf{r}k (\textbf{ShadowRefiner}) via Fast Fourier Transformer. Specifically, the Shadow Removal module in our method aims to establish effective mappings between shadow-affected and shadow-free images via spatial and frequency representation learning. To mitigate the pixel misalignment and further improve the image quality, we propose a novel Fast-Fourier Attention based Transformer (FFAT) architecture, where an innovative attention mechanism is designed for meticulous refinement. Our method wins the championship in the Perceptual Track and achieves the second best performance in the Fidelity Track of NTIRE 2024 Image Shadow Removal Challenge. Besides, comprehensive experiment result also demonstrate the compelling effectiveness of our proposed method. The code is publicly available: \url{https://github.com/movingforward100/Shadow_R}. 
\end{abstract}

%% file: section/1_intro.tex
\section{Introduction}
\vspace{-1mm}
\label{sec:intro}
Shadow-affected images typically emerge under scenarios where the light source is either partially or completely blocked, leading to spatial variations in color and illumination distortions. The objective of shadow removal is to enhance the visibility within shadow regions and achieve illumination consistency across both shadow and non-shadow areas, whilst preserving the integrity of naturalistic details. Such enhancement is pivotal for improving the performance of a plethora of downstream applications such as object detection, tracking, and segmentation systems~\cite{tracking, segmentation, segmentation2, object-detection}.

Numerous traditional methodologies proposed for image shadow removal are predominantly designed around physics-based illumination models~\cite{physical1, physical2}. Despite their theoretical underpinnings, these approaches generally exhibit limited effectiveness of removing shadows from real-world, shadow-affected images. This limitation largely stems from the difficulty in establishing an accurate physical correlation between shadow areas and their unblemished counterparts, rendering these traditional techniques less effective in practical scenarios.

Recently, learning-based approaches have emerged as a formidable mainstream within the domain of shadow removal, capitalizing on the substantial modeling capability inherent in deep learning frameworks~\cite{GridDehazeNet, GridDehazeNet+, sr4, FMSNet}. These methodologies can be bifurcated into mask-based~\cite{DeshadowNet} and mask-free shadow removal strategies, contingent upon their dependence on shadow masks for guidance. Compared to the latter, mask-based shadow removal strategies not only employ pairs of shadow-affected and shadow-free images but also integrate the location information of shadow regions, either provided by benchmark datasets or generated through pre-trained mask prediction models, as learning guidance. The introduction of precise shadow location enables these models to concentrate on unraveling the complex mappings between shadow regions and their clean counterparts, thereby achieving exceptional performance on shadow removal.

Nonetheless, the dependency on mask information unveils critical challenges to mask-based methods. \textbf{Firstly}, the acquisition of precise shadow masks is notably challenging~\cite{liu2021shadow}. Accurate shadow masks provided in public datasets are typically obtained under simply scenarios, such as a single person standing in a wide-open square. In contrast, for complex scenes, as exemplified by the WSRD and WSRD+ datasets~\cite{vasluianu2023wsrd}, annotating images with appropriate masks or employing pre-trained models to predict accurate masks proves to be impractical. \textbf{Secondly}, the absence of precise shadow masks markedly undermines the performance of mask-based models, substantially hindering their applicability to complex real-world data.

Latest advancements in mask-free shadow removal methodologies often leverage generative strategies~\cite{risgan} to learn the mappings between shadow-affected and shadow-free images. However, the adoption of frequency domain analysis remains largely under-explored within the sphere of shadow removal studies. Notably, several innovative works that integrate spatial and frequency representations has shown promising results in the broader field of image restoration, such as dehazing and low-light image enhancement task~\cite{DWT-FFC_2023_CVPRW, wcdm}, suggesting a potential direction for future exploration for shadow removal.

In this paper, we introduce a novel mask-free model that integrates spatial and frequency domain representations for image shadow removal, which achieves compelling performance on NTIRE 2024 Image Shadow Removal Challenge~\cite{vasluianu2024ntire_isr}, as illustrated in Fig.~\ref{firstfigure}. Specifically, we propose a \textbf{Shadow} Removal and \textbf{Refine}ment architecture, termed \textbf{ShadowRefiner}, with two specific modules: Shadow Removal module and Refinement module. In the \textbf{Shadow Removal} module, we design a shadow removal U-Net~\cite{unet} branch with the backbone of ConvNext~\cite{convnext} blocks. Besides, a frequency branch similar to~\cite{DWT-FFC_2023_CVPRW} equipped with high-frequency, low-frequency representations, and large receptive field, is also leveraged in our Shadow Removal module. Preliminary experiment results, however, indicate obvious pixel misalignment between the output of Shadow Removal module and the ground truth, manifesting as pronounced detail deterioration and compromised color consistency. To this end, we introduce a Fast-Fourier Attention based Transformer (FFAT) as the \textbf{Refinement} module, distinguished by its innovative attention mechanism, which significantly enhances the model's capacity to remove shadows while producing results that are simultaneously high in fidelity and visually appealing.

Our contributions are three-folds:

$\diamond$ We introduce an innovative mask-free shadow removal approach that initially clear shadow via spatial and frequency representation learning and further refined by our proposed frequency attention based transformer architecture .

$\diamond$ To mitigate the pixel misalignment, we introduce a Fast Fourier Transformer network endowed with a novel frequency attention mechanism, achieving superior performance on recovering texture details and maintaining color consistency.

$\diamond$ Extensive experiments across multiple shadow removal benchmarks, as well as the highly competitive outcomes achieved in the NTIRE 2024 Image Shadow Removal Challenge (ranking first and second in the Perceptual Track and Fidelity Track, respectively), underscore the remarkable performance of our proposed \textbf{ShadowRefiner} model.

%% file: section/figure/architecture.tex
\begin{figure*}[ht]
    \setlength{\abovecaptionskip}{2mm}
    \centering
    \begin{overpic}[width=0.99\textwidth]{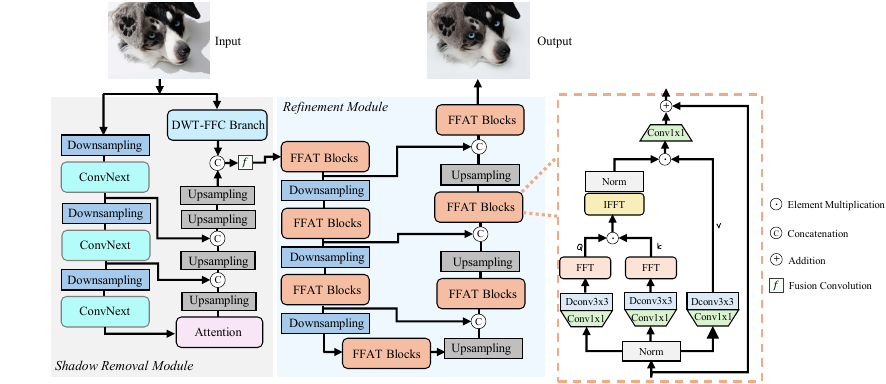}
    \end{overpic}
    \caption{The overall architecture of our model. In the Shadow Removal module, besides the DWT-FFC branch proposed in~\cite{DWT-FFC_2023_CVPRW, ancuti2023ntire}, we design a ConvNext-based U-Net architecture with $7\times7$ depth-wise convolution in each resolution. In the Refinement module, we design a new attention mechanism (Fast-Foruier Attention, FFA) different from common attention operations in transformers to further enhance texture details.} 
    \label{fig_frame}
\end{figure*}

%% file: section/2_related.tex
\section{Related Work}
\label{sec:related}

\Paragraph{Mask-based Image Shadow Removal} Mask-based methods can be categorized into two main types: feature-based and deep learning-based. Feature-based methods rely on techniques such as color constancy~\cite{colorconstancy}, texture analysis and edge detection~\cite{texture1,texture2}. Gryka~\etal~\cite{learning1} first introduce a learning-based method using a supervised regression algorithm to automatically remove umbra and penumbra shadows. Wang~\etal~\cite{stcgan} propose ST-CGAN, an end-to-end framework that integrates shadow detection and removal using two stacked Conditional Generative Adversarial Networks (CGANs). Bao~\etal~\cite{S2Net} propose S2Net that emphasizes semantic guidance and refinement for image integrity. This method uses shadow masks to guide shadow removal, with semantic-guided blocks transferring data from non-shadow to shadow areas, effectively eliminating shadows while preserving clean regions. He~\etal~\cite{Mask-ShadowNet} design Mask-ShadowNet, which ensures global illumination consistency through Masked Adaptive Instance Normalization (MAdaIN) and adaptively refines features using aligner modules. Additionally, Fu~\etal~\cite{fusionnet} introduce FusionNet, which generates fusion weight maps to eliminate shadow traces further using a boundary-aware RefineNet. However, these methods heavily rely on the accuracy of the input shadow masks. The complexity and variability of real-world scenarios could pose challenges in generating precise shadow masks, potentially affecting the performance of these methods in practical applications.

\Paragraph{Mask-free Image Shadow Removal} Fan~\etal~\cite{free1} introduce an end-to-end deep convolutional neural network consisting of an encoder-decoder network for predicting the shadow scale factor and a small refinement network for enhancing edge details. Chen~\etal~\cite{chen2021canet} design a CANet which utilizes a two-stage process for shadow removal, employing a Contextual Patch Matching (CPM) module to identify matching pairs between shadow and non-shadow patches and a Contextual Feature Transfer (CFT) mechanism to transfer contextual information, effectively eliminating shadow influence. Vasluianu~\etal~\cite{vasluianu2024image} introduce Ambient Lighting Normalization (ALN) to improve image restoration under complex lighting and propose IFBlend that enhances images by maximizing Image-Frequency joint entropy without relying on shadow localization. Liu~\etal~\cite{liu2024recasting} propose a shadow-aware decomposition network to separate illumination and reflectance layers, followed by a bilateral correction network for lighting adjustment and texture restoration.

\Paragraph{Transformer-based Image Restoration} Transformer based networks, usually adopt self-attention mechanisms to understand the relationships between different components and demonstrate high superiority in handling long dependencies, have shown state-of-the-art performance on image restoration. SwinIR~\cite{SWinIR}, a famous backbone for image restoration, is designed based on several residual Swin Transformer~\cite{swintran} blocks. With the backbone of Vision Transformer~\cite{vit}, DehazeFormer~\cite{transformer_dehazing} is proposed for dehazing task. Recently, a lightweight transformer architecture~\cite{retinexformer} is proposed for low-light image enhancement based on Retinex theory.

%% file: section/3_methods.tex
\section{Methods}
\label{sec:method}
\input{section/Tab_latex/tab_compare}

\subsection{ConvNext based U-Net for Shadow Removal} In order to achieve satisfactory shadow removal performance, powerful deep-learning networks are pivotal to extracting important features from shadow-affected images and modeling the mapping from shadow-affected and clean images. In this work, we introduce a ConvNext-based U-Net architecture, where multi-scale ConvNext blocks function as strong encoders for robust latent feature learning.

As shown in Fig.~\ref{fig_frame}, the ConvNext-based U-Net serves as the primary component in the Shadow Removal module, and the DWT-FFC branch~\cite{DWT-FFC_2023_CVPRW, wei2024dehazedct} is incorporated as an auxiliary branch. The contributions of each branch is provided in Sec.~\ref{sec_ex_abla}. 

Specifically, our ConvNext-based U-Net encompasses three downsampling layers and each downsampling operation is followed by several ConvNext blocks to feature extraction. Given a latent feature $\mbf{F}_{in}$, the ConvNext block first adopts a $7\times 7$ depthwise convolution, which functions similar to the self-attention mechanism in Transformers. Then the Layer Normalization (LN) is applied before two $1\times1$ convolutions, which are equivalent to MLP block in Transformer. Besides, only one GELU function is leveraged between two $1\times1$ convolutions inspired by the fact that Transformer MLP block incorporates only one activation function.

For the decoding process, the latent feature is aggregated using one attention block~\cite{ffanet} and several upsampling operations are utilized to recover the latent feature to their original resolution. In each upsampling layer, there are one pixel-shuffle operation and one attention block. Besides, encoder features are transferred to the decoding process via skip-connection. 

In Stage I, we only optimizing the Shadow Removal module and the training objective is shown as below:
\begin{equation}
 {L}_{loss}={L}_{1} + {\alpha}{L}_{SSIM}+{\beta}{L}_{Percep}+{\gamma}{L}_{adv}
\label{eq_loss} 
\end{equation}
where ${L}_{1}$, ${L}_{SSIM}$ and ${L}_{Percep}$ represent the $L1$ loss, MS-SSIM loss~\cite{DWT-FFC_2023_CVPRW}, and perceptual loss~\cite{VGG-16}, respectively. In addition, we leverage the discriminator proposed in ~\cite{GAN} (not provided in Fig.~\ref{fig_frame}) to calculate the adversarial loss (${L}_{adv}$). $\alpha$, $\beta$, and $\gamma$ are set to $0.2$, $0.01$, and $0.0005$ for the optimization.

\input{section/figure/ISTD/ISTD}

\subsection{Fast-Fourier Attention Transformer based Refinement}
Early experimental results suggest that our proposed ConvNext-based U-Net can effectively remove shadows, but distinct shadow contours remain, as illustrated in Fig.~\ref{fig_abla_vis}. In order to further refine image details and maintain color consistency, we introduce an efficient transformer architecture with a novel frequency attention mechanism, as shown in Fig.~\ref{fig_frame}. Similar to~\cite{Restormer}, our Refinement module adopts a encoder-decoder architecture, and our proposed FFAT blocks are utilized at each resolution level in both encoding and decoding process.  

Given a latent feature $\mbf{F}$, FFAT block first utilize $1\times 1$ point-wise convolution and $3\times 3$ depth-wise convolution to generate three features: $\mbf{Q, K, V}$. Instead of directly leveraging these features for attention calculation, we apply Fast Fourier Transform to $\mbf{Q, K}$ to calculate their frequency correlation $\mbf{A}_{F}$ based on their frequency domain representations ($\mathcal{F}(\mbf{Q})$, $\mathcal{F}(\mbf{K})$) by:
\begin{equation}
\mbf{A}_{F} = \mathcal{F}(\mbf{Q})\mathcal{F}(\mbf{K})',
\label{eq_attnetion}
\end{equation}
where $\mathcal{F}(\cdot)$ represents the FFT process and $(\cdot)'$ denotes the transpose operation. Then, the spatial correlation between $\mbf{Q}$ and $\mbf{K}$ can be obtained by the inverse FFT operation ($\mathcal{F}^{-1}(\cdot)$)  and a layer normalization ($\mathcal{LN}$). Finally, the aggregated feature $\mbf{F}_A$ and the final output of our FFAT block $\mbf{F}_{out}$ can be estimated as:
\begin{equation}
\begin{split}
&\mbf{F}_A = \mathcal{LN} ( \mathcal{F}^{-1} (\mbf{A}_{F} ) ) \mbf{V}\\
&\mbf{F}_{out} = \mathrm{Conv}_{1\times 1} (\mbf{F}_A ) + \mbf{F}.
\label{eq_FFAT}
\end{split}
\end{equation}

Compared to the global attention mechanism in Transformer and other attention strategy~\cite{AttentionLut}, our FFAT blocks can not only effectively capture long-dependencies, but also demonstrate stronger representation learning with high efficiency. This advantage stems from the frequency domain feature learning and frequency attention calculation process. To train our FFAT-based Refinement module, we first fix the Shadow Removal module and remove the adversarial loss in Eq.~\ref{eq_loss} for optimization.

%% file: section/Tab_latex/tab_compare.tex
\setlength\tabcolsep{3pt}
\begin{table*}[ht]
\setlength{\abovecaptionskip}{2mm}
\centering

        \scalebox{1}{
        \begin{tabular}{cc|ccc|ccc|ccc}
        \hline
        \multirow{2}{*}{\begin{tabular}{c}
            \textbf{ Methods}
        \end{tabular}} &\multirow{2}{*}{\begin{tabular}{c}
            \textbf{ Mask-free}
        \end{tabular}} &\multicolumn{3}{c|}{ISTD~\cite{stcgan}} &\multicolumn{3}{c|}{ISTD+~\cite{le2019shadow}} &\multicolumn{3}{c}{WSRD+~\cite{vasluianu2023wsrd}}  \\\cline{3-11}
        & &PSNR$\uparrow$ &SSIM$\uparrow$ &LPIPS$\downarrow$ &PSNR$\uparrow$ &SSIM$\uparrow$ &LPIPS$\downarrow$   &PSNR$\uparrow$ &SSIM$\uparrow$ &LPIPS$\downarrow$\\ 
        \hline
        DHAN~\cite{dhan} &No &24.86 &0.919 &0.0535 &27.88 &0.917 &0.0529  &22.39 &0.796 &0.1049 \\
        BMNet~\cite{bmnet} &No &29.02 &0.923 &0.0529 &31.85 &0.932 &0.0432 &24.75 &0.816 &0.0948\\
        FusionNet~\cite{fusionnet} &No &25.84 &0.712 &0.3196 &27.61 &0.725 &0.3123 &21.66 &0.752 &0.1227 \\
        SADC~\cite{SADC} &No &29.22 &0.928 &0.0403 &--- &--- &--- &--- &--- &--- \\
        ShadowFormer~\cite{shadowformer} &No &\blue{30.47} &\blue{0.928} &\blue{0.0418} &\blue{32.78} &\blue{0.934} &\blue{0.0385} &\blue{25.44} &\blue{0.820} &\blue{0.0898} \\
        \hline
        DCShadowNet~\cite{dcshadow} &Yes &24.02 &0.677 &0.4423 &25.50 &0.694 &0.4237 &21.62 &0.593 &0.4744 \\
        Refusion~\cite{refusion} &Yes &25.13 &0.871 &0.0571 &26.28. &0.887 &0.0437 &22.32 &0.738 & 0.0937 \\
        \hline
        \textbf{ShadowRefiner (Ours)} &Yes &\red{28.75} &\red{0.916} &\red{0.0521} &\red{31.03} &\red{0.928} &\red{0.0426} &\red{26.04} &\red{0.827} &\red{0.0854} \\
        \hline
        \end{tabular}
    }
    
\caption{Quantitative comparisons with SOTA methods. Our ShadowRefiner significantly outperforms other mask-free methods across three benchmarks. Compared to mask-based methods, our ShadowRefiner achieves comparable or even better performance (WSRD+ dataset). [Key: \red{Best performance among mask-free models}, \blue{Best performance among mask-based methods}]}
\label{table_quant_compar}
\end{table*}

\setlength\tabcolsep{6pt}

%% file: section/figure/ISTD/ISTD.tex
\begin{figure*}[t]
    \setlength{\abovecaptionskip}{1mm}
    \centering
    \includegraphics[width=1\textwidth]{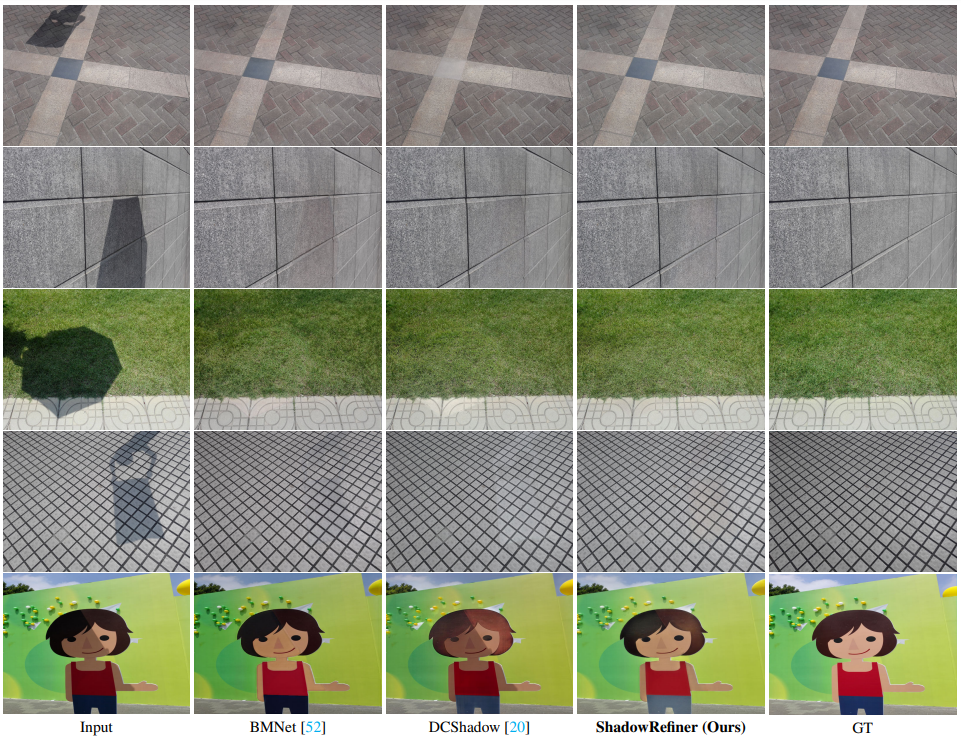}
    \caption{Visual comparisons on the ISTD dataset~\cite{stcgan}. Compared to other methods, our ShadowRefiner successfully remove shadows without incorporating artifacts.}
    \label{ISTD}
\end{figure*}

%% file: section/4_experiments.tex
\input{section/figure/ISTD+/ISTD+}

\input{section/Tab_latex/order}

\Section{Experiments}
\label{sec_ex}

\subsection{Datasets and Implementation Details}
\Paragraph{Datasets} We evaluate our proposed method on WSRD+~\cite{vasluianu2023wsrd}, ISTD~\cite{stcgan}, and ISTD+~\cite{le2019shadow} datasets. WSRD+ dataset, as the enhanced version of the WSRD dataset with improved pixel-alignment, is used as the benchmark dataset for NTIRE 2024 Image Shadow Removal Challenge. This dataset consists of 1200 high-resolution image pairs. The training set, validation set, and test set are split in proportions of 10:1:1, and we train our model on the training set and evaluate the performance on the validation set. ISTD dataset contains 1870 image triplets obtained from 135 distinct scenarios, of which 1330 are assigned for training and the remaining 540 are for testing. ISTD+ dataset is a color-adjusted version of ISTD and it has the same number and structure as the ISTD dataset.

\Paragraph{Implementation Details} One RTX 2080Ti GPU is used to execute the two-stage training of our method. For data augmentation, we implement random cropping of patches with dimensions of \(384 \times 384\), combined with random rotations of \(90\), \(180\), or \(270\) degrees, as well as vertical and horizontal flipping. The Adam optimizer with the default hyper-parameters, where \(\beta_1\) and \(\beta_2\) are set to \(0.9\) and \(0.999\) respectively, is utilized for optimization. In Stage I, only Shadow Removal module is optimized and the learning rate is initially set to \(1\times10^{-4}\) and is gradually reduced to \(6.25\times10^{-6}\). In Stage II, we adopt a constant learning of \(1\times10^{-5}\) to simultaneously update the Shadow Removal module and Refinement module.

\Paragraph{Evaluation Metrics} To comprehensively evaluate the performance of various shadow removal methods, three metrics are adopted for quantitative comparison: The Peak Signal to Noise Ratio (PSNR), the Structural Similarity Index (SSIM)~\cite{SSIM}, and Learned Perceptual Image Patch Similarity (LPIPS)~\cite{LPIPS} to assess the pixel-level accuracy and perceptual quality.

\subsection{Comparison with State-of-the-Art Methods} 

We compare our proposed method with several State-of-the-art (SOTA) algorithms. Specifically, several mask-free methods including Refusion~\cite{refusion}, DCShadowNet~\cite{dcshadow} and recent proposed mask-based approaches including ShadowFormer~\cite{shadowformer}, SADC~\cite{SADC} are adopted for comparison. 

\Paragraph{Quantitative Results} As documented in Tab.~\ref{table_quant_compar}, our ShadowRefiner demonstrates superior performances compared to other mask-free methods across three datasets. On the ISTD dataset~\cite{stcgan}, there is a marked improvement of 3.62 dB in PSNR, a 0.045 increase in SSIM, and a 0.005 decline in LPIPS. On the ISTD+ dataset~\cite{le2019shadow}, we observe a 4.75 dB increase in PSNR and a 0.041 increase in SSIM. Besides, our ShadowRefiner yields a 3.72 dB increase in PSNR, a 0.089 increase in SSIM, and a 0.008 decline in LPIPS on the WSRD+ dataset~\cite{vasluianu2024ntire_isr}. Furthermore, compared to mask-based methods, which require shadow mask information and naturally outperform mask-based models, our mask-free ShadowRefiner is capable to generate comparable results on ISTD and ISTD+ datasets. Moreover, on WSRD+ dataset where precise mask image is unavailable, though a mask prediction method proposed in~\cite{dhan} is utilized to assist these mask-based models, our ShadowRefiner achieves superior performance than the best mask-based approach (ShadowFormer~\cite{shadowformer}), demonstrating that our ShadowRefiner can work effectively in complex scenarios. 

\Paragraph{Qualitative Comparisons} Visual comparisons on ISTD dataset, ISTD+ dataset and WSRD+ dataset are reported in Fig.~\ref{ISTD},~\ref{ISTD+}, and ~\ref{wsrd}, respectively. Obviously, the results of our method closely match GT images in both color and detail preservation. For instance, in Fig.~\ref{ISTD} row 3, our ShadowRefiner removes the shadow without introducing artifacts or discoloration, issues commonly seen with other methods. In Fig.~\ref{ISTD+} row 2, our method skillfully adjusts the shadow areas on the pavement, effectively lightening the shadows to blend with the sunlit parts without distorting the underlying patterns or hues. Moreover, for images featuring toys and colorful objects shown in Fig.~\ref{wsrd}, our ShadowRefiner restores the vivid colors and intricate patterns that are obscured by shadows in the input images.

\input{section/figure/wsrd+/wsrd+}

\subsection{Performance on NTIRE 2024 Shadow Removal Challenge (Fidelity and Perceptual Track)} According to the challenge report~\cite{vasluianu2024ntire_isr}, our model is the \textbf{first place} of \textbf{Perceptual Track} and the \textbf{second place} of \textbf{Fidelity Track} with the highest SSIM (0.832) and  Mean Opinion Scor (MOS, 7.750)  and competitive PSNR (24.58), demonstrating advanced performance of our method on shadow removal task. We also report the result of our method for the official validation and test data used in NTIRE 2024 Shadow Removal Challenge as Fig.~\ref{firstfigure} and Fig.~\ref{wsrd}. We can see that our ShadowRefiner achieves notably satisfying shadow removal effect in pictures of different scenarios. For example, the shadow in the first image in Fig.~\ref{firstfigure} is completely eliminated, retaining the integrity of the scattered toy blocks in their original layout. Additionally, the color consistency between toys initially positioned in shadowed areas and those under direct illumination attests to the high fidelity in color reproduction. The restoration of the final image depicting a dog conveys a sense of uniform, soft lighting across the entire subject, further exemplifying ShadowRefiner's capability to enhance illumination homogeneity.

%% file: section/figure/ISTD+/ISTD+.tex
\begin{figure*}[t]
    \setlength{\abovecaptionskip}{1mm}
    \centering
    \includegraphics[width=1\textwidth]{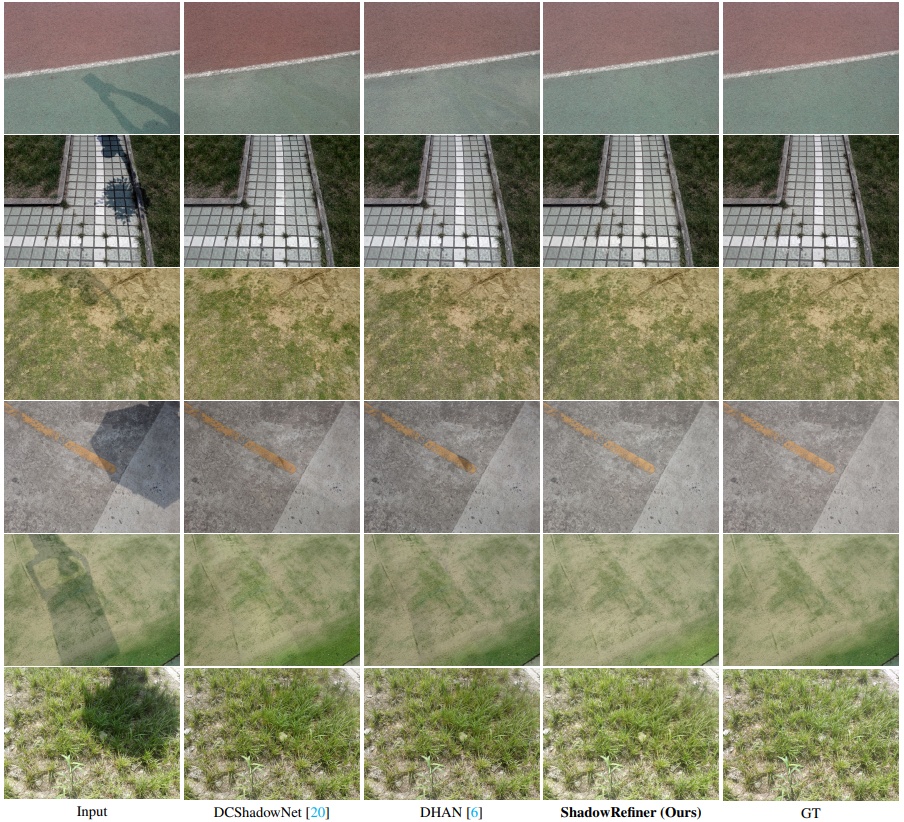}
    \caption{Visual comparisons on the ISTD+ dataset\cite{le2019shadow}. Obviously, our ShadowRefiner excels in maintaining color consistency and recovering structure details}
    \label{ISTD+}
\end{figure*}

%% file: section/Tab_latex/order.tex
\begin{table*}[!h]
    \setlength{\abovecaptionskip}{2mm}
    \centering
    \scalebox{1}{
    \begin{tabular}{c|cccc|c}
    \hline
        
    \textbf{Team} & PSNR$\uparrow$ & SSIM$\uparrow$ & LPIPS$\downarrow$ & MOS$\uparrow$ & Fianl Rank$\downarrow$  \\ 
    \hline
        \textbf{Shadow\_R (Ours)}                           &24.578 &\textbf{\red{0.832}} &0.098 &\textbf{\red{7.750}} &\textbf{\red{1}}  \\
        LVGroup\_HFUT                           &24.232 &0.821 &\textbf{\red{0.082}} &\blue{7.519} &2  \\
        USTC\_ShadowTitan                      &24.042 &0.827 &0.104 &7.444 &3  \\
        ShadowTech Innovators                      &\textbf{\red{24.810}} &0.832 &0.111 &7.438 &4   \\
        GGBond                      &23.050 &0.809 &\blue{0.089} &7.400 &5   \\
        PSU Team                      &22.219 &0.731 &0.132 &7.400 &6   \\
        LUMOS                      &\blue{24.783} &\blue{0.832} &0.110 &7.163 &7   \\
        IIM\_TTI                      &22.955 &0.806 &0.093 &7.160 &8   \\
        AiRiA\_Vision                      &21.902 &0.689 &0.238 &6.825 &9  \\
        HKUST-VIP\_Lab\_01                      &22.284 &0.788 &0.135226084 &6.619 &10  \\
   
        \hline
        \end{tabular}
    }
\caption{Final ranking (top 10 teams) of Perceptual Track in NTIRE 2024 Shadow Removal Challenge. Our solution achieves the best performance among all 19 submitted solutions. [Key: \textbf{\red{Best}}, \blue{Second Best}, $\uparrow (\downarrow)$: The larger (smaller) represents the better performance].}    
\label{order}
\end{table*}

%% file: section/figure/wsrd+/wsrd+.tex
\begin{figure*}[t]
    \setlength{\abovecaptionskip}{2mm}
    \centering
    \includegraphics[width=1\textwidth]{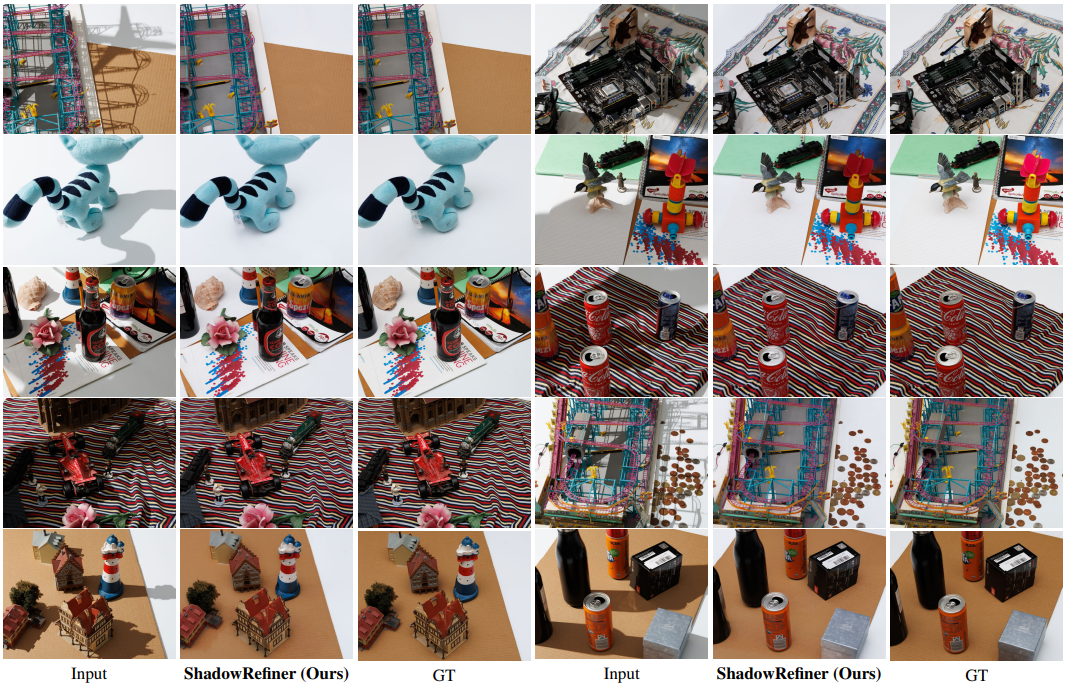}
    \caption{Our results on the validation set of WSRD+ dataset~\cite{vasluianu2023wsrd}. Our method performs well on color and detail recovery.}
    \label{wsrd}
    \vspace{2mm}
\end{figure*}

%% file: section/4_experiment_ablation.tex
\input{section/Tab_latex/tab_abla}
\input{section/figure/ablation/abla_vis}

\SubSection{Ablation Study}
\label{sec_ex_abla}

In this section, we conduct several ablation experiments on WSRD+ dataset~\cite{vasluianu2023wsrd}. 

\Paragraph {Importance of FFAT based Refinement module} To study the contribution of the Refinement module and our proposed FFAT blocks, we first remove the Refinement module to figure out its contribution for the outstanding performance of our method provided in Tab.~\ref{table_quant_compar}. The quantitative result is reported in Tab.~\ref{tab_abla}, which demonstrates removing Refinement module leads to obviously degraded performance. Then, we replace our Refinement module with Restormer module~\cite{Restormer} and we notice a marked decrease in PSNR (0.36 dB $\downarrow$) and SSIM (0.009 $\downarrow$) and a discernible increase of LPIPS (0.0024 $\uparrow$). Visual comparisons provided in Fig.~\ref{fig_abla_vis} also demonstrate our Refinement module can help generate more visually appealing results compared to the Restormer module. This ablation experiment underscore the importance of our proposed Refinement module for satisfactory shadow removal performance. 
  
\Paragraph {Contributions of ConvNext-based U-Net} Based on the configuration for Tab.~\ref{tab_abla} row 2, we implement several further adaptations to illustrate the effectiveness of ConvNext-based U-Net and the DWT-FFC branch leveraged in the Shadow Removal module. By separately comparing row 4 and row 5 to row 2 in Tab.~\ref{tab_abla}, we can conclude that adopting a two-branch architecture in the Shadow Removal module help achieve more pleasant performance than single branch. Moreover, compared to the DWT-FFC branch, the enhanced performance of the ConvNext-based U-Net architecture highlights its predominant role within the Shadow Removal module.

%% file: section/Tab_latex/tab_abla.tex
\setlength\tabcolsep{3.2pt}
\begin{table}[t]
    \setlength{\abovecaptionskip}{2mm}
    \centering
    \scalebox{1}{
    \begin{tabular}{c|ccc}
    \hline
        
    \textbf{Configurations} &PSNR$\uparrow$  &SSIM$\uparrow$ &LPIPS$\downarrow$ \\ 
    \hline
    \textbf{ShadowRefiner (Ours)} &\red{\textbf{26.04}} &\red{\textbf{0.827}} &\red{\textbf{0.0854}} \\
    w/o Refinement module &25.54 &0.816 &0.0886 \\
    w/ Restormer module~\cite{Restormer} &\blue{25.68} &\blue{0.818} &\blue{0.0878} \\
    only ConvNext-based U-Net &25.29 &0.810 &0.0914 \\    
    only DWT-FFC branch &23.36 &0.791 &0.1016\\
    \hline
    \end{tabular}
    }    
\caption{The ablation results on WRSD+ dataset. Each component in our ShadowRefiner help achieve competitive performance on shadow removal task, and our proposed Refinement module performs better than Restormer.}
\label{tab_abla}
\vspace{-3mm}
\end{table}
\setlength\tabcolsep{6pt}

%% file: section/figure/ablation/abla_vis.tex
\begin{figure}[ht]
    \setlength{\abovecaptionskip}{2mm}
    \centering
    \includegraphics[width=0.47\textwidth]{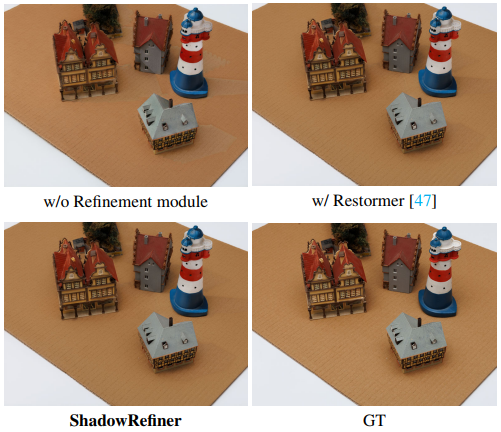}
    \caption{Visual ablation comparisons on the WSRD+ dataset.}
    \label{fig_abla_vis}
    \vspace{-2mm}
\end{figure}

%% file: section/5_conclusions.tex
\section{Conclusion}
\label{sec:conclu}

In this paper, we propose \textbf{ShadowRefiner}, a novel mask-free model for shadow removal task. Specifically, the Shadow Removal module with ConvNext-based U-Net is firstly introduced to extract spatial and frequency representations and effectively learning the mapping between shadow-affected and clean images. Then, we design a novel transformer module based on Fast Fourier Attention Transformer to enhance color and structure consistency. Extensive experiments demonstrate ShadowRefiner significantly outperforms the current mask-free methods and its capacity is comparable to mask-based shadow removal approaches. Furthermore, our method wins the \textbf{championship} in the Perceptual Track and \textbf{ranks second} in the Fidelity Track of NTIRE 2024 Image Shadow Removal Challenge.